\documentclass[a4paper]{article}
\usepackage{multirow}

\usepackage{INTERSPEECH2022}

\title{The THUEE System Description for the IARPA OpenASR21 Challenge}
\name{Jing Zhao, Haoyu Wang, Jinpeng Li, Shuzhou Chai, Guan-Bo Wang, Guoguo Chen, Wei-Qiang~Zhang$^*$
\thanks{$^*$Corresponding author}
\thanks{This work was supported by the National Natural Science Foundation of China under Grant No. U1836219.}}
\address{
Beijing National Research Center for Information Science and Technology \\
  Department of Electronic Engineering,
Tsinghua University, Beijing 100084, China}
\email{wqzhang@tsinghua.edu.cn}

\begin{document}

\maketitle
\begin{abstract}
  This paper describes the THUEE team's speech recognition system for the IARPA Open Automatic Speech Recognition Challenge (OpenASR21), with further experiment explorations. We achieve outstanding results under both the Constrained and Constrained-plus training conditions. For the Constrained training condition, we construct our basic
  ASR system based on the standard hybrid
  architecture. To alleviate the Out-Of-Vocabulary (OOV) problem, we extend the pronunciation lexicon using Grapheme-to-Phoneme (G2P) techniques for both OOV and potential new words. Standard acoustic model structures such as CNN-TDNN-F and CNN-TDNN-F-A
  are adopted. In addition, multiple data augmentation techniques are applied.
  For the Constrained-plus training condition, we use the self-supervised learning framework wav2vec2.0. We experiment with various fine-tuning techniques with the Connectionist Temporal Classification (CTC) criterion on top of the publicly available pre-trained model XLSR-53. We find that the frontend feature extractor plays an important role when  applying the wav2vec2.0 pre-trained model to the encoder-decoder based CTC/Attention ASR architecture. Extra improvements can be achieved by using the CTC model finetuned in the target language as the frontend feature extractor.

\end{abstract}

\section{Introduction}

The OpenASR21 Challenge\footnote{https://sat.nist.gov/openasr21} is the third open challenge created out of the Intelligence Advanced Research Projects Activity (IARPA) Machine Translation for English Retrieval of Information in Any Language (MATERIAL) program\footnote{ https://www.iarpa.gov/index.php/research-programs/material}. The goal of the OpenASR21 is to assess the state of the art of ASR technologies for low-resource languages.
For most of the languages in the world, there are no applicable ASR systems due to the lack of high-quality annotated speech data. It is challenging to build a strong ASR system with limited speech data, transcripts as well as lexicons. The capabilities tested in the open challenges are expected to ultimately support the MATERIAL task of effective triage and analysis of large volumes of data in a variety of less-studied languages\footnote{https://www.nist.gov/document/openasr21-challenge-evaluation-plan}.

We describe our ASR systems in detail to demonstrate our solutions for challenge in the Constrained and Constrained-plus training conditions respectively. Post evaluation experiments for the Constrained-plus training condition are also provided for future references.

For the Constrained training condition,
we focus on hybrid acoustic models with CNN-TDNN-F and CNN-TDNN-F-A network structures\cite{zhaoj2021}.
To alleviate the OOV problem, which significantly impacts the speech recognition performance in low-resource conditions, we extend our lexicon using G2P techniques. OOV words from the training data and their corresponding pronunciations generated by G2P techniques are first added to the base lexicon. The lexicon is then extended by generating potential new words in the target language using the method described in \cite{trmal2014keyword}. Recurrent neural network language models are applied by performing lattice rescoring forward and backward successively \cite{liu2014efficient}. To further improve the robustness, we combine various data augmentation methods \cite{KoPPK15,ParkCZCZCL19,SnyderCP15} during the training process, and we train three different systems for each language for a final system combination.


For the Constrained-plus training condition,
unsupervised pre-trained models are allowed in addition to the provided 10-hour dataset. We achieve significant improvements with extra unlabeled speech data from the BABEL program\footnote{https://www.iarpa.gov/index.php/research-programs/babel} and unsupervised pre-trained wav2vec2.0 \cite{BaevskiZMA20} models compared to the Constrained training condition. We use two finetuning methods to build our systems, and different systems are also processed by different Speech Activity Detection (SAD) methods. Results from those systems are fused to yield the single best result.
Besides, we also explore various ways of incorporating the pre-trained wav2vec2.0 models in the encoder-decoder based CTC/Attention ASR system \cite{WatanabeHKHH17}.

The rest of the paper is organized as follows. Sec.\ref{sec:openasr21} introduces the OpenASR21 challenge briefly, and Sec.\ref{sec:constrained} presents our systems for the Constrained training condition. We describe our ASR systems for the Constrained-plus training condition in Sec.\ref{sec:constrained-plus}, together with our post evaluation experiments with the unsupervised pre-trained models. Finally, conclusions are drawn in Sec.\ref{sec:conclusion}.


\section{The OpenASR21 Challenge}
\label{sec:openasr21}

The OpenASR21 challenge is an open challenge to evaluate the ASR performance under low-resource language constraints.
There are 15 languages in total, three of which have additional evaluation datasets scored using Case-Sensitive Criteria (CSS). Basic information of the 15 languages is shown in Tab. \ref{tab:15lang_data}. For each language, three different training conditions are set: Constrained, Constrained-plus, and
Unconstrained. The main difference among the three conditions is the restriction on speech data and models. For the Constrained training condition, the only permissible resource is a 10-hour dataset provided by the National Institute of Standards and Technology (NIST) for the language being processed. For Constrained-plus, unsupervised pre-trained models are allowed as well. There is no limitation for Unconstrained condition as long as the data are publicly available. Text data are not restricted for all three conditions.

\begin{table}[h]
\caption{The 15 languages in OpenASR21 challenge. }
\begin{center}
\begin{tabular}{llll}
\toprule
\textbf{Language}  & \textbf{Code}  & \textbf{Dataset}   & \textbf{Scoring}\\
\midrule
{Amharic} & am & BABEL & CIS\\
{Cantonese} & yue & BABEL  & CIS\\
Farsi & fa & MATERIAL & CIS\\
Georgian & ka & BABEL & CIS\\
{Guarani} & gu & BABEL  & CIS \\
{Javanese} & jv & BABEL & CIS \\
{Kazakh} & kk & BABEL & CIS\&CSS\\
{Kurmanji-Kurdish} & ku & BABEL & CIS\\
{Mongolian} & mn& BABEL & CIS\\
{Pashto} & ps& BABEL  & CIS\\
Somali & so & MATERIAL & CIS\\
{Swahili} & sw &  BABEL & CIS\&CSS\\
{Tagalog} & tl & BABEL & CIS\&CSS \\
{Tamil} & ta & BABEL  & CIS\\
{Vietnamese} & vi& BABEL & CIS\\
\bottomrule
\end{tabular}
\label{tab:15lang_data}
\end{center}
\end{table}

For Case-Insensitive Scoring (CIS) evaluation, the provided Build datasets consist of only one data genre, conversational telephone speech, which has two separate channels. The CSS datasets also include news broadcast and topical broadcast. The sampling rate of speech data is 8 kHz, 44.1 kHz or 48 kHz.

\section{The Constrained Training Condition}
\label{sec:constrained}

We build hybrid Deep Neural Networks-Hidden Markov Model (DNN-HMM) ASR systems for the Constrained training condition with Kaldi speech recognition toolkit \cite{Povey11}.
The important parts of our system, including pronunciation lexicons, acoustic model, data augmentation and pre-and-post processing, are described one by one in the following sections.




\subsection{Pronunciation Lexicon}
It is essential to deal with OOV problem for low-resource languages. We use the pronunciation lexicons in BABEL program for the languages from BABEL program. In addition, the text data in Babel datasets are utilized for training language model as shown in Tab. \ref{tab:babel_data}. For datasets from MATERIAL program and datasets using CSS evaluation criterion, OOV problems are more serious.
Therefore, we adopt G2P models to generate approximate pronunciations for the OOV words in these datasets, and then append the generated items to the original lexicons. The G2P models are from LanguageNet \cite{hasegawa2020grapheme} which provides Phonetisaurus \cite{Phonetisaurus} FST G2P models for around 150 languages.

Furthermore, we perform lexicon expansion for all the 18 datasets, which generates new items composed of word, corresponding pronunciation and probability. In the process, we first produce 12 million phone sequences based on a 3-gram language model trained with items in the original lexicon. After eliminating the phone sequences already existing in the lexicon, we keep 1 million items with highest probabilities to obtain the corresponding spellings by G2P methods. Finally, we expand the original lexicon with 1 million new entries.

\subsection{Acoustic Model}


The CNN-TDNN-F network \cite{GeorgescuCB19} is adopted as the neural network acoustic model, which combines Convolution Neural Network (CNN), Factored Time Delay Neural Network (TDNN-F) \cite{povey2018semi}. We also employ CNN-TDNN-F-A architecture \cite{zhaoj2021} to build multiple systems with diversity, which introduces self-attention mechanism \cite{VaswaniSPUJGKP17,PoveyHGLK18} to CNN-TDNN-F network.
For both two networks, there are 6 convolution blocks at the beginning and 11 TDNN-F blocks in total with the hidden dimension of $768$ and a bottleneck dimension of $160$.
As for CNN-TDNN-F-A architecture, the self-attention block is the third layer from the output end, which is composed of an affine component, an attention nonlinearity component, and a ReLU nonlinearity component followed by batch normalization \cite{PoveyHGLK18}.
The acoustic models are trained with high-resolution MFCC features concatenated with i-vector features \cite{saon2013speaker} for speaker adaptation. For Cantonese, Georgian, Tagalog and Tamil, pitch features are appended as well.
The acoustic models are trained with chain/chain2 component of Kaldi toolkit which adopts LF-MMI criterion. 
We resample all speech data to 8 kHz.
\begin{table}[h]
\caption{The extra available data from IARPA BABEL program. }
\begin{center}
\begin{tabular}{llll}
\toprule
\textbf{Lang}  & \textbf{Version}  & \textbf{\#words}  & \textbf{Dur} \\
\midrule
{am} & IARPA-babel307b-v1.0b-build & 281k & 43h\\
{yue} & IARPA-babel101b-v0.4c-build & 892k  & 141h \\
{gu} & IARPA-babel305b-v1.0c-build & 311k & 42h  \\
{jv} & IARPA-babel402b-v1.0b-build & 309k  & 45h \\
{kk} & IARPA-babel302b-v1.0a-build & 270k & 39h\\
{ku} & IARPA-babel205b-v1.0a-build & 346k & 41h\\
{mn} & IARPA-babel401b-v2.0b-build & 403k &46h \\
{ps} & IARPA-babel104b-v0.bY-build & 888k  & 78h\\
{sw} & IARPA-babel202b-v1.0d-build &  287k & 44h\\
{tl} & IARPA-babel106-v0.2g-build & 595k & 85h\\
{ta} & IARPA-babel204b-v1.1b-build & 486k  & 69h\\
{vi} & IARPA-babel107b-v0.7-build & 923k & 88h\\
\bottomrule
\end{tabular}
\label{tab:babel_data}
\end{center}
\end{table}






\subsection{Data Augmentation}

We apply several techniques for data augmentation at the same time to enhance the robustness of our ASR systems, including speed and volume perturbation \cite{KoPPK15}, SpecAugment \cite{ParkCZCZCL19} and adding noise.
By speed perturbation, we enlarge the data quantity by three times with speed factors of 0.9, 1.0 and 1.1.
 We quadruple the amount of data with MUSAN dataset \cite{SnyderCP15} by different noise type: babble, music, noise and reverberation. For reverberation, we add simulated Room Impulse Responses (RIRs) of small and middle room at the same time  \cite{Ko2017ASO}.

\subsection{Pre-and-Post Processes}
\label{sec:pre-and-post}
For the evaluation period, Speech Activity Detection (SAD) is a necessary operation to segment the audio appropriately so that we can decrease loss of useful speech clips and improve the decoding efficiency and accuracy. The SAD system follows work proposed in \cite{wang2019fusion,shi2021timestamp}.
In our SAD component, we combine a Recurrent Neural Network (RNN) and a Convolutional RNN (CRNN) to make the system more robust.
Additionally, we also utilize an subband Order Statistic Filters (OSFs) based VAD system \cite{ramirez2005effective} to increase diversity of the fused system. 


The first-pass decoding is based on Weighted Finite-State Transducers (WFST), incorporating an N-gram language model.
We also employ lattice rescoring with NN-based Language Model (LM), composed by TDNN-LSTM networks \cite{liu2014efficient}.
Furthermore, we perform backward rescoring with the same network trained on the reversed text.
For system fusion, we adopt the lattice combination method \cite{2010An}, which uses minimum Bayes risk decoding to combine results from the same SAD system. Then, we merge the combined results with the help of ROVER method \cite{fiscus1997post}.


\subsection{Results on DEV}

We evaluate our systems with the 10h development set. The results are shown in Tab. \ref{tab:results-constrained} by Word Error Rate (WER). We show 4 single systems' results and their fusion results for each language and scoring criteria.


\begin{table}[tbh]
\caption{The results (WER) of DEV under the Constrained training condition. $S1$ and $S2$ employ the CNN-TDNN-F and CNN-TDNN-F-A respectively. $S3$ performs RNNLM rescoring based on $S2$. $S4$ is the augmented with MUSAN based on $S1$ or $S2$. $Fusion$ is obtained with $S1$-$S4$ by lattice combination. }
\begin{center}
\begin{tabular}{lllllll}
\toprule
\textbf{Lang} & \textbf{Case} & \textbf{S1}  & \textbf{S2}  & \textbf{S3} & \textbf{S4} & \textbf{Fusion}  \\
\midrule
{am} & {CIS} & 37.3 & 37.7 &38.5 & 40.9 &35.3\\
{yue} & {CIS} &47.3 &47.8 &46.6 &48.7 & 45.0 \\
{fa} & {CIS} & 54.4 & 54.2 &55.0 &58.0 & 52.6\\
{ka} & {CIS} &42.8 & 42.3 & 44.3& 47.2 &40.8\\
{gu} & {CIS} & 41.0& 41.8 &41.6 & 44.9 &38.9 \\
{jv} & {CIS} &54.0 &54.5 &54.3&57.6 &51.9 \\
\multirow{2}{*}{kk} & {CIS} &  46.4 &47.9 &46.4& 52.9 &44.2 \\
 & {CSS} & 29.9 &30.6 &30.4&33.7 & 27.6\\
{ku} & {CIS} &66.1 & 66.3 &65.6&70.5 & 63.8\\
{mn} & {CIS} & 48.3& 48.6 & 48.4&52.5 &45.5 \\
{ps} & {CIS} &47.0 & 47.2 & 46.3& 51.2 & 44.1\\
{so} & {CIS} &55.9 & 56.6  &56.4&59.7 & 53.8 \\
\multirow{2}{*}{sw} & {CIS} &34.4 & 34.8 &34.5& 37.9 &32.2 \\
 & {CSS} &27.5 & 28.0 &29.8& 29.3 & 26.4\\
\multirow{2}{*}{tl} & {CIS} &42.2 & 42.9 &41.9&44.1 & 40.0\\
 & {CSS} &33.4 & 33.3 &30.9& 34.1 & 28.5 \\
{ta} & {CIS} & 62.0 & 62.9 &62.9& 65.4 & 60.2\\
{vi} & {CIS} &47.4 & 47.0 &46.4& 50.3 & 44.2\\
{Ave} & -- & \textbf{45.4}  & \textbf{45.8} &\textbf{45.6} &\textbf{48.8} & \textbf{43.1}\\
\bottomrule
\end{tabular}
\label{tab:results-constrained}
\end{center}
\end{table}

\section{The Constrained-plus Training Condition}
\label{sec:constrained-plus}

The Constrained-plus training condition follows the same training data restrictions as the Constrained training condition, but additionally allows publicly available
speech pre-trained models. Recently the self-supervised learning frameworks, such as wav2vec2.0 \cite{BaevskiZMA20} and HuBERT \cite{hsu2021hubert},
have achieved promising progress in speech recognition. The self-supervised models are able to transfer across languages and be applicable with small amount of labeled data \cite{cpc_modified,xlsr}.
We build our pipeline based on the wav2vec2.0 framework \cite{BaevskiZMA20} for its outstanding performances on ASR downstream task. The wav2vec2.0 model is composed of a multi-layer convolutional feature encoder, a context network with Transformer architecture, and a quantization module to discretize the output of the feature encoder to a finite set of speech representations via product quantization \cite{BaevskiZMA20}.
In our ASR systems under the Constrained-plus training condition, we finetune the pre-trained model in two stages to take advantage of unlabeled speech data from target language. In addition, we further explore the application of wav2vec2.0 pre-trained model in End-to-End (E2E) ASR systems. Details are described in the following sections respectively.

\begin{table}[tbh]
\caption{The results (WER) of DEV
under the Constrained-plus training condition, $yue$ is shown by Char Error Rate (CER). $FT$ is to finetune the XLSR-53 model directly while $FT2$ continues training the XLSR-53 model first, which is introduced as the 2-stage finetuning method in Sec.\ref{subsec:finetuning}. (w/ oov) means the texts used during finetuning do not remove OOVs. $Frontend$ refers to the CTC/Attention ASR system introduced in Sec.\ref{sec:post}. }
\begin{center}
\begin{tabular}{llccccc}
\toprule
\multirow{2}{*}{\textbf{Lang}}  & \multirow{2}{*}{\textbf{Case}} & \textbf{FT}  & \textbf{FT2}  & \multicolumn{2}{|c|}{\textbf{FT2(w/ oov)}} & \textbf{Frontend}\\
\cline{3-7}
& & \multicolumn{3}{c|}{4-gram LM} & \multicolumn{2}{c}{w/o LM} \\
\midrule
{am} & {CIS} &44.0 & 40.8 & 38.6 &44.9 & 43.4 \\
{yue} & {CIS} & 37.0 & 37.1 &  36.6 &36.7 & 37.1  \\
{fa} & {CIS} & 46.3 &45.9 &47.4 & 49.1& 43.5 \\
{ka} & {CIS} & 44.7& 45.1 & 47.3 &50.7 & 45.5 \\
{gu} & {CIS} & 46.2 & 42.5  & 41.3 & 46.4& 43.5 \\
{jv} & {CIS} & 53.6 & 51.0  & 49.8 & 56.9& 53.0 \\
\multirow{2}{*}{kk} & {CIS} &  46.5& 43.9 & 42.1 & 48.3& 45.6 \\
 & {CSS} & 50.9& 48.8& 38.3 & -- & --\\
{ku} & {CIS} &62.5  & 60.1 & 59.5 &66.9 & 64.4\\
{mn} & {CIS} & 47.9 & 44.7 & 43.9 &51.4 & 46.7\\
{ps} & {CIS} & 45.2 & 41.8  & 37.9 & 47.1 & 44.6\\
{so} & {CIS} & 53.9 & 53.1 & 55.8 & 62.6 & 56.7 \\
\multirow{2}{*}{sw} & {CIS} & 40.5 & 36.8 & 34.6 & 39.1 & 37.4 \\
 & {CSS} & 48.6&45.4 & 38.1 & -- & -- \\
\multirow{2}{*}{tl} & {CIS} & 45.0& 41.5  &39.5 &46.8 & 43.3\\
 & {CSS} & 43.6& 42.1& 34.3 & --& --\\
{ta} & {CIS} & 64.3 & 61.8  & 60.0 & 69.1 & 64.6 \\
{vi} & {CIS} & 40.2 & 35.9 & 36.6 &  44.0 & 39.3\\
\multicolumn{2}{c}{Average}  & \textbf{47.8}& \textbf{45.5}  & \textbf{43.4} & \textbf{50.7}  & \textbf{47.2} \\
\bottomrule
\end{tabular}
\label{tab:results-constrained-plus}
\end{center}
\end{table}

\subsection{Finetuning CTC Model}
\label{subsec:finetuning}

For the Constrained-plus training condition, we conduct our experiments on the fairseq toolkit\footnote{https://github.com/pytorch/fairseq}. Since there are 15 different languages in the challenge, we utilize an open-source multilingual pre-trained model \cite{xlsr} XLSR-53\footnote{https://dl.fbaipublicfiles.com/fairseq/wav2vec/xlsr\_53\_56k.pt} as backbone, which is trained with 56k hours audio data in 53 different languages from 3 datasets (Multilingual LibriSpeech \cite{Pratap20}, CommonVoice \cite{ArdilaBDKMHMSTW20}, BABEL \cite{GalesKRR14}).
In the basic workflow, we finetune the pre-trained XLSR-53 model with the labeled 10h data of the target language by adding a linear classifier on top of the model to optimize the CTC loss  \cite{ctc}. The provided 10h development set is used to validate during finetuning. For the first 10k updates only the output classifier is trained, after which the Transformer is also updated. The feature encoder is not trained during finetuning. For model compatibility, all the audio data used are up or down sampled to 16kHz.

We improve the basic finetuning methods in our systems. Multilingual pre-trained models are the experts in acquiring
universal characteristics of multiple languages. However,  they may not contain enough unique information for a specific language since the language identities are not utilized during pre-training. Therefore, we further train the XLSR-53 model by self-supervised learning with unlabeled speech of the target language before finetuning. At this stage, the whole model is optimized by contrastive loss augmented by a codebook diversity loss \cite{BaevskiZMA20}. Next, we finetune the language-specific XLSR-53 model with the labeled 10h data as  the basic finetuning method.
The speech data used for unsupervised training are from the Build dataset in BABEL program of the corresponding language being processed, which have been shown in Tab. \ref{tab:babel_data}. For the three languages not listed, Farsi, Georgian and Somali, we utilize the same 10h data in two stages.




We put character as tokenization unit for languages with romanized spelling, such as Tagalog and Swahili,
while grapheme is utilized for languages without romanized spelling.
The token set also includes a word boundary token. For CIS datasets, we transfer all uppercase letters into lowercase letters while keeping the originl case for CSS datasets. Besides, we deal with the transcript texts in two ways after removing all the speech aspects such as mispronunciations and non-speech aspects.
One is keeping all words in transcripts while the other is filtering out OOV words with the provided pronunciation lexicons.


For experiment settings of unsupervised training, the masking probability of a time-step is 65\%.
We optimize with Adam optimizer, warming up the learning rate for the first 32000 updates to a peak of $1 $×$ 10^{-3}$ and then linearly decaying it. The maximum number of updates is 100k.
During finetuning, we mask 75\% of the time-steps and 25\% of the channels. The learning rate follows a tri-state rate schedule where it is warmed up for the first 10\% of updates, held at $1 $×$ 10^{-3}$ for the next 40\% and then linearly decayed for the remainder.

The results on DEV are displayed in Table~\ref{tab:results-constrained-plus}. We decode with a 4-gram language model, of which the weight is set to 1 with beam 5. By comparing the average results, the proposed 2-stage finetuning method is validated to be effective, which improves the performance by 9.2\% relatively compared with basic finetuning. Note that the XLSR-53 model has already utilized the same speech data from BABEL of 8 languages ($yue$, $kk$, $ku$, $ps$, $sw$, $tl$, $ta$, $vi$) out of the 15 languages during pre-training, which verifies that the self-training stage with speech data of target language is necessary and it is critical to apply the multilingual model to a specific language. For most languages, the text processing method of keeping OOVs benefits the performances.
For evaluations of the challenge, we perform SAD on the datasets
with two different methods mentioned in Sec.\ref{sec:pre-and-post}. The final result is fused from results of different segment methods and different fine-tuning ways by ROVER \cite{fiscus1997post}. We do not present the detailed fusion results due to the limited space.




\subsection{Post Evaluations with wav2vec2.0 Pre-trained Model}
\label{sec:post}

\begin{figure}[tbh]
  \centering
  \includegraphics[width=0.9\linewidth]{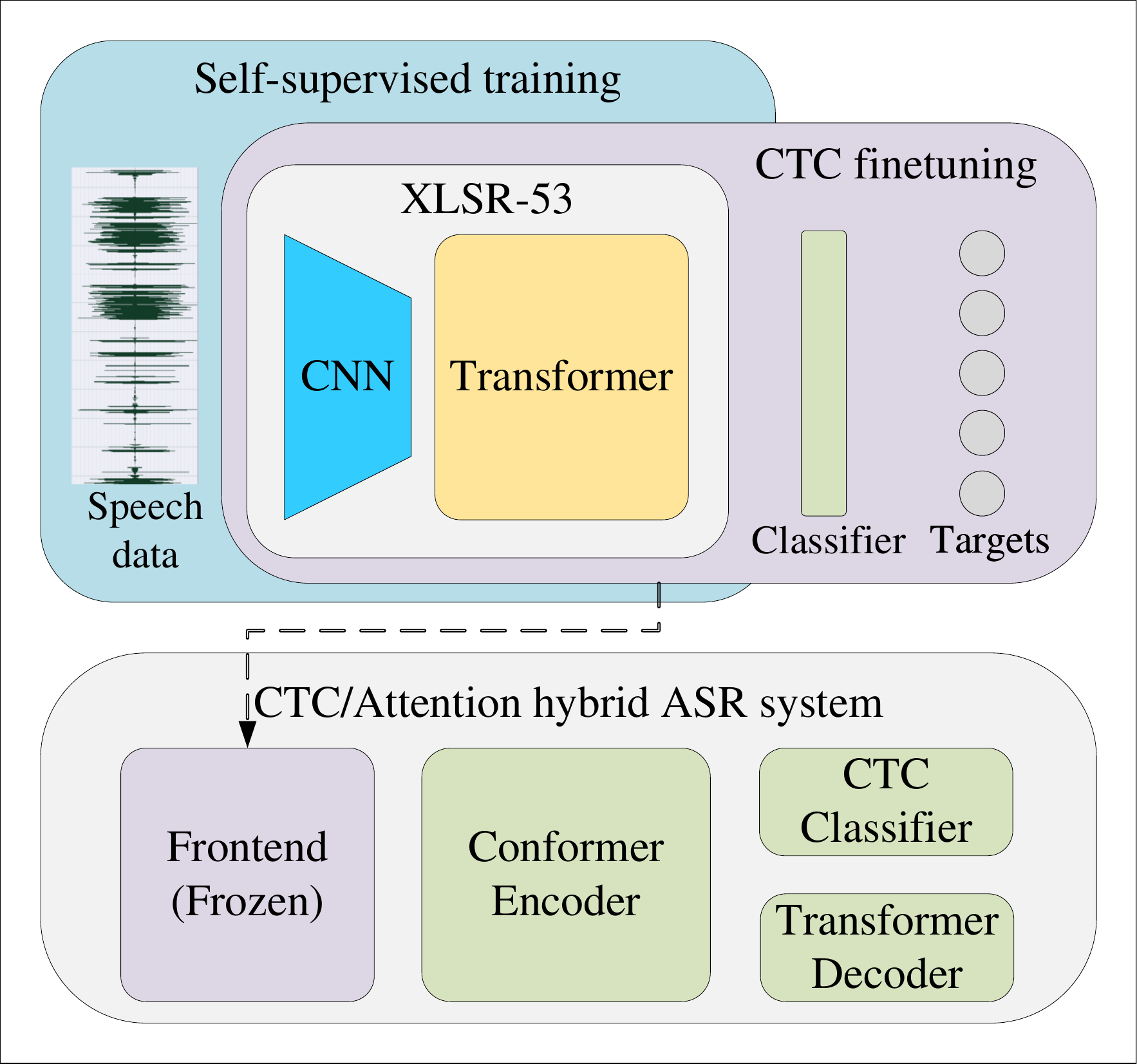}
  \caption{ASR system based on XLSR-53 pre-trained model}
  \label{fig:ft2-conformer}
\end{figure}

Recently the E2E systems based on Transformer or Conformer architecture \cite{speech_transformer,conformer} have achieved promising performances in speech recognition. However, they do not benefit low-resource languages directly owing to the data-driven property of E2E systems. Therefore, it is reasonable to combine the self-supervised model based on large quantity of speech data and data-hungry E2E ASR system for low-resource languages. We treat the pre-trained wav2vec2.0 model as a frontend module to take advantage of the effective representations extracted from the pre-trained model. To further improve the ASR performance on low-resource languages, we adopt the CTC/Attention hybrid E2E system structure \cite{WatanabeHKHH17} enhanced by wav2vec2.0 model under the  Constrained-plus training condition, which is illustrated in Figure~\ref{fig:ft2-conformer}. We utilize the 2-stage finetuned XLSR-53 model for the target language as the system frontend.

In CTC/Attention hybrid system, we adopt Unigram method \cite{unigram} for tokenization with a vocabulary size of 500, except for Cantonese, which is over 2,000 decided by the number of characters appeared.
500 is a suitable size for most languages according to our ablation experiments.
For system architecture, the encoder is composed by 12 Conformer blocks, and the decoder is composed by 6 Transformer blocks. There are 8 attention heads with 512 dimensions and the number of hidden-units in feed forward layer are 2048. The 1024-dimension embeddings extracted from pre-trained model are first projected to 80-dimension representations before input to encoder. For optimization objective, the weight of CTC loss is set to 0.3.
The experiments are performed with ESPnet \cite{watanabe2018espnet}.

We perform experiments on the 15 languages as shown in Table~\ref{tab:results-constrained-plus}. The right two columns compare the performances of 2-stage finetuned wav2vev2.0 system and CTC/Attention E2E system with wav2vec2.0 frontend, both of which are obtained without language model in order to eliminate the interference of different language models. The CTC/Attention hybrid system obtains lower WER by 6.9\% on average. Therefore, it is more competitive to apply the finetuned wav2vec2.0 model to the Conformer-based E2E ASR system as frontend.

\section{Conclusion}
\label{sec:conclusion}

We describe the THUEE ASR systems of 15 low-resource languages in detail under both the Constrained and Constrained-plus training condition in OpenASR21 challenge. We have achieved outstanding results
, especially for Constrain-plus condition. The effective combination of classical methods under Constrained condition is a valuable case for the researchers on low-resource languages. For Constrained-plus condition, we obtain remarkable performances with the 2-stage finetuning methods based on wav2vec2.0 pre-trained model. In addition, we further explore the application of pre-trained model to CTC/Attention hybrid ASR system, which verifies the effectiveness of the frontend utilization.

\bibliographystyle{IEEEtran}

\bibliography{THUEE}


\end{document}